\pdfoutput=1

\documentclass[11pt]{article}

\usepackage{emnlp2021}

\usepackage{times}
\usepackage{latexsym}

\usepackage[T1]{fontenc}

\usepackage[utf8]{inputenc}

\usepackage{microtype}
\usepackage{amsmath}
\usepackage{amssymb}
\usepackage{pifont}
\usepackage{graphicx}
\usepackage{graphics}
\usepackage{hyperref}
\usepackage{url}
\usepackage{multirow}
\usepackage{multicol}
\usepackage{subfigure}
\usepackage{algorithm}
\usepackage{algorithmic}
\usepackage{color}
\usepackage{booktabs}
\usepackage{arydshln}
\usepackage{stfloats}

\newcommand{\method}{Generate \& Rank}
\renewcommand\footnotemark{}

\title{\method: A Multi-task Framework\\ for Math Word Problems}

\author{Jianhao Shen\textsuperscript{1$^\dag$}\thanks{$^\dag$ This work is done when Jianhao Shen is an intern at Huawei Noah’s Ark Lab},  Yichun Yin\textsuperscript{2}, Lin Li\textsuperscript{3}, Lifeng Shang\textsuperscript{2},\\ {\bf Xin Jiang}\textsuperscript{2},  {\bf Ming Zhang}\textsuperscript{1*}\thanks{*Corresponding author}, {\bf Qun Liu}\textsuperscript{2} \\
\textsuperscript{1}Department of Computer Science, School of EECS, Peking University \\
\textsuperscript{2}Huawei Noah's Ark Lab\\
\textsuperscript{3}Huawei HiSilicon\\
\{jhshen, mzhang\_cs\}@pku.edu.cn \\
\{yinyichun, lilin29, shang.lifeng, jiang.xin, qun.liu\}@huawei.com
}

\begin{document}
\maketitle

\begin{abstract}
Math word problem (MWP) is a challenging and critical task in natural language processing. Many recent studies formalize MWP as a generation task and have adopted sequence-to-sequence models to transform problem descriptions to mathematical expressions. However, mathematical expressions are prone to minor mistakes while the generation objective does not explicitly handle such mistakes. To address this limitation, we devise a new ranking task for MWP and propose \method, a multi-task framework based on a generative pre-trained language model. By joint training with generation and ranking, the model learns from its own mistakes and is able to distinguish between correct and incorrect expressions. Meanwhile, we perform tree-based disturbance specially designed for MWP and an online update to boost the ranker. We demonstrate the effectiveness of our proposed method on the benchmark and the results show that our method consistently outperforms baselines in all datasets. Particularly, in the classical Math23k, our method is {\bf 7\%} (78.4\% $\rightarrow$ {\bf 85.4\%}) higher than the state-of-the-art\footnote{Code will be available soon.}.
\end{abstract}
\section{Introduction}
Solving math word problems (MWP)~\cite{bobrow_natural_1964} is an important and fundamental task in natural language processing (NLP), which requires to provide a solution expression given a mathematical problem description, as illustrated in Table~\ref{tab:mwp}. Many recent studies formalize MWP as a generation task and commonly adopt LSTM-based sequence-to-sequence (Seq2Seq) models~\cite{wang-etal-2017-deep-neural, wang_translating_2018, xie_goal-driven_2019}, where problem texts are source sequences, mathematical expressions are target sequences and the model learns the mapping from source texts to target expressions. These studies have proposed numerous advanced techniques to improve the MWP solver, but their performance is still unsatisfactory yet.

\begin{table}[!t]
    \centering
    \resizebox{0.48\textwidth}{!}{
    \begin{tabular}{|c|p{0.4\textwidth}|}
    \hline
         \multicolumn{2}{|c|}{{\bf Original MWP}} \\
    \hline
         Problem & A project is completed in 25 days by 12 workers.  If it takes 20 days to complete, how many workers will it take? \\
    \hline
         Solution & {\it 25 * 12 / 20}  \\ 
    \hline
         \multicolumn{2}{|c|}{{\bf Number-mapped MWP}} \\
    \hline
         Problem & A project is completed in {\it NUM0} days by {\it NUM1} workers.  If it takes {\it NUM2} days to complete, how many workers will it take? \\
    \hline
         Solution & {\it NUM0 * NUM1 / NUM2}  \\ 
    \hline
    \end{tabular}}
    \caption{An example of MWP, where numbers are usually mapped to special tokens, such as {\it Num0/1/2}.}
    \label{tab:mwp}
\end{table}


We argue that it is not sufficient to model MWP as only a generation task, because there is a significant difference between mathematical expressions and natural language sequences: one minor mistake in a mathematical expression will change the whole semantic thus lead to a wrong answer, whereas natural language is more robust to such minor mistakes. The objective function of the generation task is to maximize generation likelihood on ground-truth expressions, which does not have an explicit strategy to make the model learn to distinguish between ground-truth and expressions that have minor mistakes. In addition, previous works~\cite{liu_tree-structured_2019, xie_goal-driven_2019, zhang_graph--tree_2020} find that the performance of generation models degrades fast as the expression gets longer.

To handle the above problems, we propose \method, a multi-task framework for MWP, which introduces a new ranker to explicitly distinguish between correct and incorrect expressions. Specifically, our framework includes two modules: a generator and a ranker. The former is designed to generate candidate expressions given a problem text and the latter aims to rank the candidate expressions. They are built based on an encoder-decoder model and are jointly trained with generation loss and ranking loss. In this work, we build our model based on BART~\cite{lewis_bart:_2020}, a widely used pre-trained language model that achieves SOTA performance on various sequence-to-sequence tasks~\cite{ahmad2020summarization, liu_multilingual_2020}. During multi-task training, expressions produced by the generator are used to construct an expression bank and train the ranker, in which way the model can learn from its own mistakes. To construct more informative candidates for the ranker, we specially design tree-based disturbance for MWP. We also introduce an online update mechanism to generate a new set of candidate expressions at each training epoch. The overall training procedure is in an iterative manner, in which the ranker and generator continue to enhance each other.

To evaluate the effectiveness of the proposed model, we conduct extensive experiments on the datasets of Math23K~\cite{wang-etal-2017-deep-neural} and MAWPS~\cite{koncel-kedziorski_mawps:_2016}. The results show that our model outperforms typical baselines.
Particularly, we obtain an improvement of 7\% in the Math23K dataset that is extensively studied. Moreover, we do ablation study and model analysis, which shows that (1) joint training improves the performance of the generator and ranker over separate training; (2) both strategies of constructing candidate expressions and online updating are important to the success of the ranker. We also find that with the ranker, our model achieves a large improvement in generation of long expressions.

The contributions of our work are two-fold: (1) We propose \method, a new multi-task framework to train a pre-trained language model for math word problem solving. To construct informative candidate expressions for the ranker, we propose two effective generation methods and also introduce an online update strategy. (2)  Experiments show that our proposed model consistently outperforms the state-of-the-art models and achieves a significant improvement on the Math23K dataset.

\section{Preliminaries}
\subsection{Math Word Problem}
A math word problem {\it P} is a sequence of word tokens and numeric values, which typically describes a partial quantitative state of a world and some updates or relationships among quantities, then asks a question about an unknown quantity. The solution {\it S} to the question is a mathematical expression that consists of math operators and numbers. In solving a math word problem, we usually do not care about the specific number of a quantity, so the numbers in problems and solution expressions are mapped to special tokens \textit{NUM\#i} according to their orders in the problem text. Table \ref{tab:mwp} gives an example of an original math word problem and the corresponding number-mapped problem.

\begin{figure*}[h]
    \centering
    \includegraphics[width=0.9\textwidth]{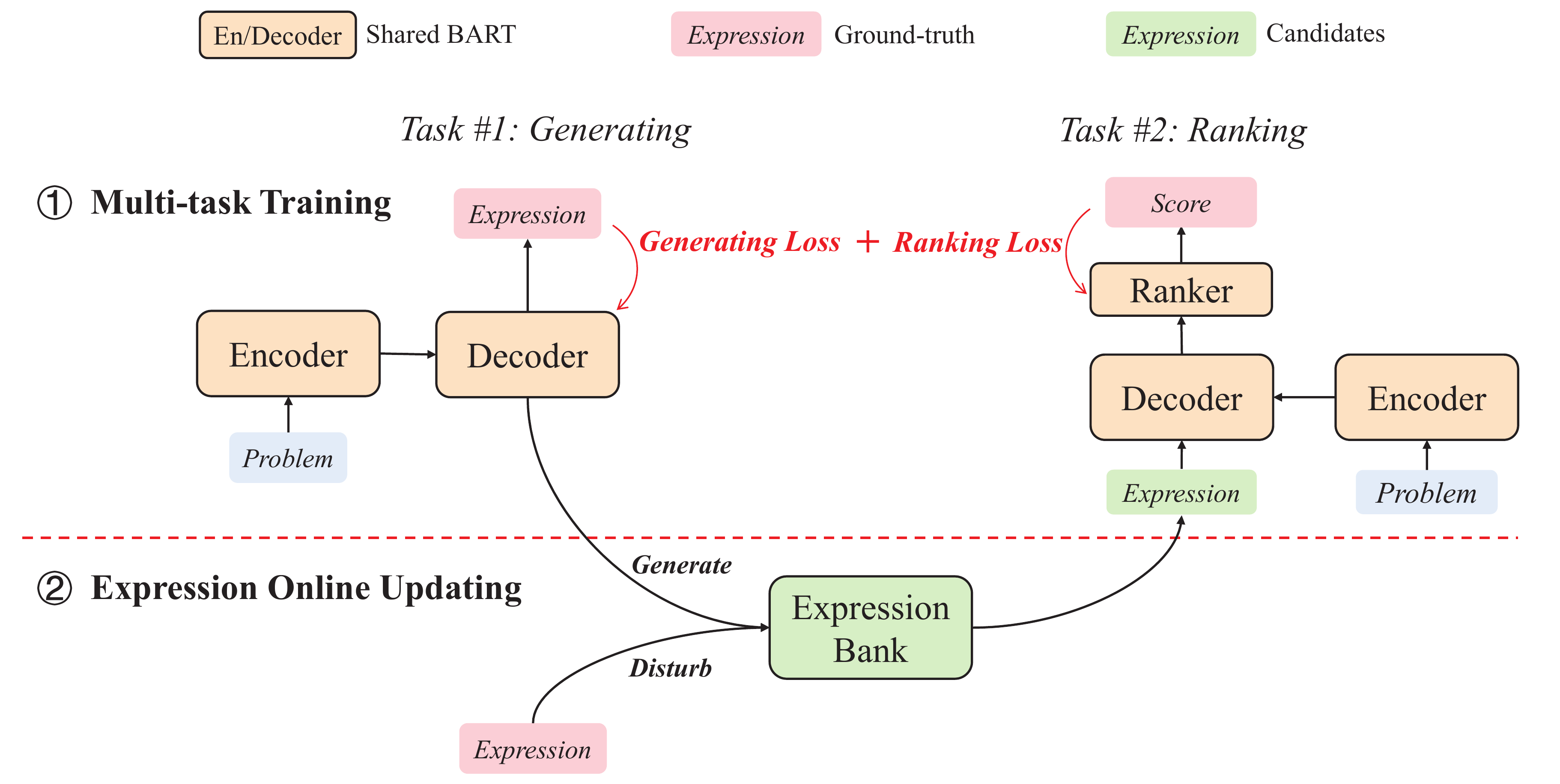}
    \caption{Our proposed \method\ framework for BART-based MWP solver. The model consists of a generator and a ranker. They share BART encoder and decoder, and are jointly trained with generating loss and ranking loss. We construct an expression bank for training the ranker with expressions produced by the generator and ones obtained by tree-based disturbance. The expression bank is updated every epoch so that the model can constantly learn from new informative examples.}
    \label{fig:model}
\end{figure*}

\vspace{2mm}
\subsection{BART}
BART is a widely-used pre-trained language model. It follows a standard encoder-decoder structure using Transformer layers~\cite{Vaswani_attention} and is pre-trained with text denoising tasks. The pre-trained BART can be fine-tuned for tasks of sequence classification and generation.

\vspace{2mm}
\noindent {\bf Transformer-based Encoder-Decoder.}
BART uses an encoder-decoder structure that is the mainstream architecture for sequence-to-sequence tasks. The encoder adopts the bidirectional self-attention to map an input sequence of tokens $P = (x_1, x_2, \dots, x_n)$ to a sequence of continuous representations $\bf{R} = (\bf{r}_1, \bf{r}_2, \dots, \bf{r}_n)$. The BART encoder is composed of multiple Transformer layers, each consists of a multi-head self-attention (MHA) module and a fully connected feed-forward (FFN) module. We denote the mapping function of the BART encoder as follows: 
\begin{equation}
    ({\bf r_1}, {\bf r_2}, \dots, {\bf r_n}) = {\rm BART}_{\rm Enc}(x_1, x_2, \dots, x_n)
\end{equation}

The BART decoder also consists of multiple Transformer layers. Besides MHA and FFN modules, the decoder layer adds another multi-head attention over the output of the encoder. The decoder takes in one token $s_i$ at a time, and gives an output state based on the output of the encoder and previous tokens in the decoder input. This output state is then fed into a linear transformation followed by a softmax function to get the predicted next-token probabilities. This one-step decoding process is denoted as follows:
\begin{align}
    P(*) &= {\rm softmax}({\bf d}_i {\bf W}+ {\bf b}) \\
    {\bf d}_i &= {\rm BART}_{\rm Dec}(\mathbf{R}; s_0, s_1, \dots, s_{i-1}),
\end{align}
where $s_0$ is a special \textit{[bos]} token indicating the start of decoding, and $\mathbf{R}$ is the output of encoder.

\vspace{2mm}
\noindent {\bf BART Pre-training.}
BART is pre-trained by the tasks of recovering a corrupted document to original one. The input to BART is corrupted in two ways: (1) a number of text spans are replaced with a single \textit{[MASK]} token; (2) sentences in a document are shuffled in a random order. The objective of BART pre-training is to minimize the cross-entropy loss between the decoder's generation probabilities and the ground-truth of original document.


\section{Methodology}

We propose \method, a BART-based multi-task framework for math word problems. Our model consists of a generator and a ranker, which share a BART model and are jointly trained with a generating task and ranking task. The objective of generating is to generate expressions given a math word problem. We also add a ranking task so that the model can select a correct expression from a set of candidates. We construct an expression bank to provide training examples for the ranker. Figure \ref{fig:model} shows our proposed framework and we introduce details for each task and the whole framework in the following sections.

\subsection{Multi-task Training}
\noindent {\bf Task \#1: Generating.}
We first formulate the math word problem as a sequence-to-sequence task, in which BART is trained to generate solution expressions given a math word problem. Following the fine-tuning strategy of BART~\cite{lewis_bart:_2020}, we take problem text, a sequence of tokens $P = (x_1, x_2, \dots, x_n)$, as input to BART encoder, and minimize negative log-likelihood of the solution expression $S = (s_1, s_2, \dots, s_m)$,
\begin{equation}\label{eq:gen_loss}
    J_{\rm GEN} = \frac{1}{|\mathcal{D}|}\sum_{(P,S) \in \mathcal{D}} -\log \Pr(S|P),
\end{equation}
where the conditional probability is decomposed in an auto-regressive way as:
\begin{align}
    \Pr(S|P) &= \prod_{i=1}^m \Pr(s_i|P, S_{j<i}) \\
    \Pr(*|P, S_{j<i}) &= {\rm softmax}({\bf d}_i{\bf W} + {\bf b}) \\
    {\bf d}_i &= {\rm BART}_{\rm Dec}({\bf R}; S_{j<i}) \\
    \mathbf{R} &= {\rm BART}_{\rm Enc}(P).
\end{align}
Additionally, we add two special tokens $s_1 =$\textit{[bos]} and $s_m = $\textit{[eos]} to indicate the start and end symbols of decoding sequences. 




\vspace{2mm}
\noindent{\bf Task \#2: Ranking.}
Through generating, we obtain many candidate solution expressions. To decide which expression is a correct solution to the problem, we propose a ranking task which is essentially a task of sequence pair classification. Given pairs of problems and candidate expressions, the ranker chooses the expression with highest ranking score as the final solution to the problem. Specifically, we add an MLP classifier on top of the final layer hidden state of the last decoder token. The last decoder token is always a special \textit{[eos]} token and its corresponding hidden state can attend to all token representations of problem text and expression. Same as the generation task, we feed the problem text into the encoder and expression into the decoder, obtaining sequence representations. The last decoder representation is then taken as input to the classifier for ranking score prediction:
\begin{align}
\Pr(\cdot|P,S) &= {\rm softmax}({\bf d}_{m+1}’) \\
{\bf d}_{m+1}’ &= {\rm tanh}({\bf d}_{m+1}{\bf W}_1 + {\bf b}_1){\bf W}_2 + {\bf b}_2 \\
    {\bf d}_{m+1} &= {\rm BART}_{\rm Dec}({\bf R}; S) \label{eq:verify},
\end{align}

\noindent where ${\bf R}$ is the output of the encoder, $S$ is the expression token sequence, ${\bf d}_{m+1}$ is the decoder representation of the last token, and ${\bf W}_{1|2}$ and ${\bf b}_{1|2}$ are trainable parameters.
The training objective of the ranker is cross-entropy between classifier output and correct labels,
\begin{equation}
\begin{aligned}
        J_{\rm RANK} = &-\frac{1}{|\mathcal{D}^+ \cup \mathcal{D}^-|}\bigg[\sum_{(P,S) \in \mathcal{D}^+} \log\Pr(1|P,S) \\
        &+\sum_{(P,S)\in \mathcal{D}^-} \log\Pr(0|P,S) \bigg]
\end{aligned}
\end{equation}
\noindent where $\mathcal{D}^+$ and $\mathcal{D}^-$ are sets of positive and negative examples, respectively. We introduce how to generate negative examples in the next section.

\vspace{2mm}
\noindent{\bf Optimization Objective.}
We train the model on the joint loss of two tasks together:
\begin{equation}\label{eq:joint_loss}
    J = J_{\rm GEN} + J_{\rm RANK}.
\end{equation} 
and the two modules share BART parameters.

\subsection{Expression Bank}
By definition, any expression that does not equal the ground-truth can serve as a negative example, but we cannot use all of them due to limited computational resources. To train the ranker efficiently, we use two different strategies, namely model-based generation and tree-based disturbance, to construct an expression bank for ranker training.

\vspace{2mm}
\noindent{\bf Model-based Generation.} The first strategy is to produce new expressions with the generator. Specifically, given a problem, we use beam search with the generator to produce top-$K$ expressions. Each expression is labeled as positive or negative depending on whether its calculation result equals the result of ground-truth.

\begin{figure}[h]
    \centering
    \includegraphics[width=0.47\textwidth]{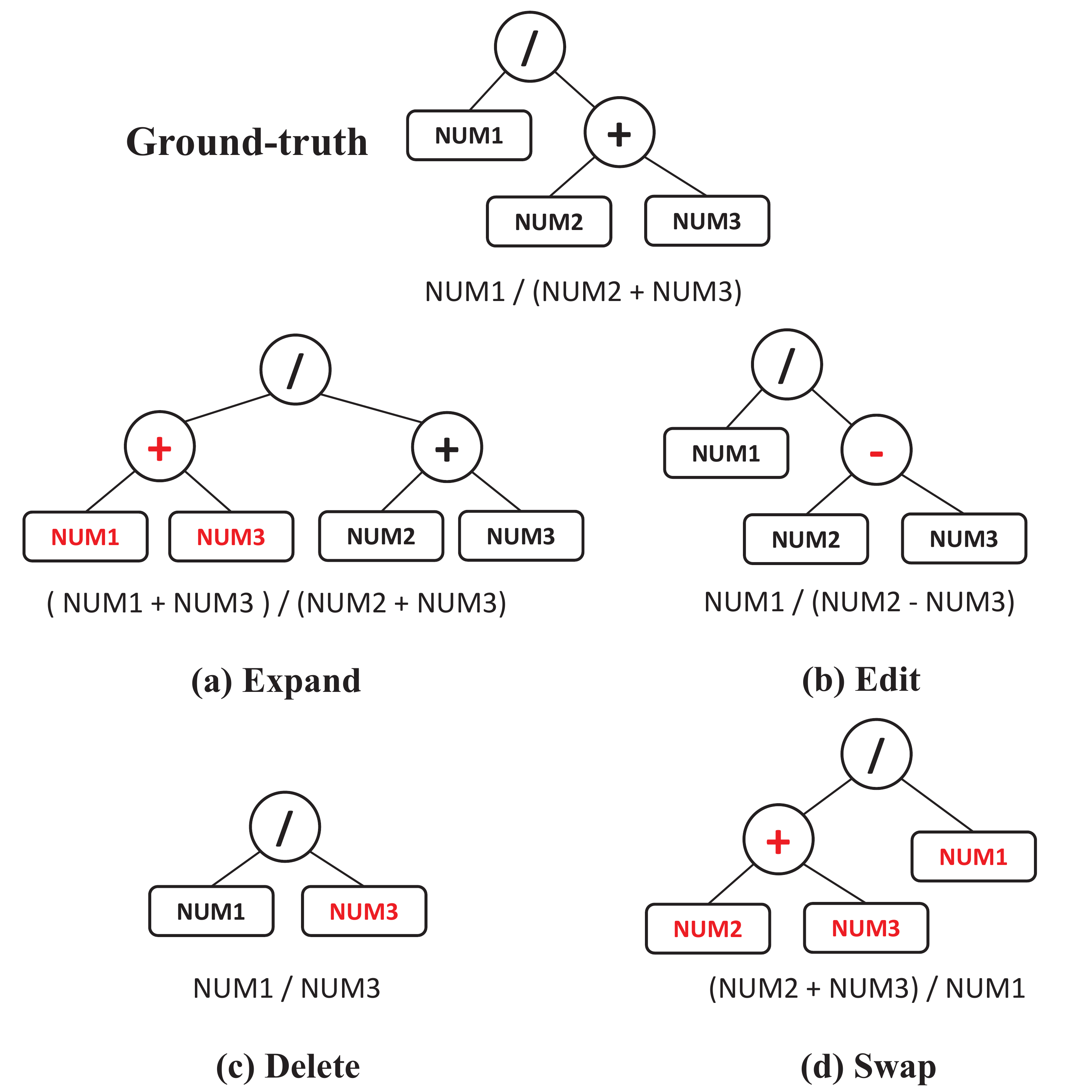}
    \caption{Overview of tree-based disturbance.}
    \label{fig:disturbance}
\end{figure}

\vspace{2mm}
\noindent{\bf Tree-based Disturbance.} Our second way to construct new  expressions is adding disturbance to ground-truth expressions. We design four kinds of disturbances which are illustrated in Figure \ref{fig:disturbance}. The ground-truth expression is first transformed to an abstract syntax tree (AST)~\cite{liu_tree-structured_2019}. Then we disturb tree nodes or sub-structures to produce new expressions in four ways: {\bf a) Expand.} A leaf node is expanded into a sub-tree with a new operation and a number. {\bf b) Edit.} A node is randomly changed to another while keeping the expression valid (i.e., a number node will be changed to another number, and an operator node to another operator). {\bf c) Delete.} Delete a leaf node and replace its father with its sibling node. {\bf d) Swap.} Swap the left and right children of an operation node.

We use the above methods to construct the expression bank. Since new expressions may also be correct (for example, swapping two operands of addition or multiplication), we compare the numerical results of newly obtained expressions with that of the ground-truth, and add them to positive or negative samples depending on the comparison. Then both positive and negative pairs are sampled from this expression bank for the multi-task training. In order to make the model learn with more informative examples, we do an online update for expression bank, which means that we use new expressions obtained by model-based generation and tree-based disturbance at each training epoch. 


\subsection{Training Procedure}
The training procedure includes multi-task training and expression online updating. We first fine-tune the pre-trained BART for the generation task ($J_{\rm GEN}$ in Eq. \ref{eq:gen_loss}). After that, we use the fine-tuned BART and tree-based disturbance to generate expressions as the training samples for the ranker. Then we do the joint training of generation and ranking. This process is performed in an iterative manner and the two modules (i.e., generator and ranker) continue to enhance each other. Meanwhile, training examples for ranking are updated after each epoch. We summarize the overall training procedure in Algorithm \ref{alg:train}.

\begin{algorithm}[tb]
\caption{Training Algorithm}
\label{alg:train}
\textbf{Input:} MWP Dataset $\mathcal{D} = \{(P,S)\}$ \\
\textbf{Parameter:} Pre-trained BART encoder and decoder parameters $\theta_e$ and $\theta_d$, random initialized ranker $\theta_v$, beam size $K$, epoch number $M$

\begin{algorithmic}[1] 
\STATE \textrm{// Fine-tune the generator}
\FOR{$epoch=1$ to $M$}
\STATE Fine-tuning BART encoder $\theta_e$ and decoder $\theta_d$ on $\mathcal{D}$ with generation loss Eq. (\ref{eq:gen_loss}).
\ENDFOR
\STATE \textrm{// Construct expression bank}
\STATE $\mathcal{D}^+ \gets \mathcal{D}$, $\mathcal{D}^- \gets \{\}$
\FOR{$(P,S) \in \mathcal{D}$}
    \STATE Generate top-$K$ expressions $\{\bar{S}_i\}$ for problem $P$ with beam search 
    \STATE Get new expressions $\{\bar{S}'_i\}$ by adding tree-based disturbance to $S$
    \STATE $\{\bar{S}_i\} \gets \{\bar{S}_i\} \cup \{\bar{S}'_i\}$
    \FOR{$\bar{S} \in \{\bar{S}_i\}$}
        \IF{ result of $\bar{S}$ equals result of $S$}
            \STATE $\mathcal{D}^+ \gets \mathcal{D}^+ \cup \{(P,\bar{S})\}$
        \ELSE
            \STATE $\mathcal{D}^- \gets \mathcal{D}^- \cup \{(P,\bar{S})\}$
        \ENDIF
    \ENDFOR
\ENDFOR
\STATE \textrm{// Joint training}
\FOR{$epoch=1$ to $M$}
    \STATE Train $\theta_e$, $\theta_d$, $\theta_v$ w.r.t. the joint loss Eq.(\ref{eq:joint_loss}) on $\mathcal{D}^+$ and $\mathcal{D}^-$ 
    \STATE Repeat lines 6-18 to reconstruct expression bank
\ENDFOR
\end{algorithmic}
\end{algorithm}

\vspace{2mm}
\subsection{Model Inference}
We perform a two-stage model inference, namely generation and ranking. Specifically, given a new problem text sequence $P$, we first pass it to the encoder to get the problem representation {\bf R}. Then we perform the beam search to generate top-$K$ expressions. These generated expressions are used as candidate solutions for the ranker. All expressions are passed to the ranker and that with the highest score is selected as the final result. 



\section{Experiment}
\subsection{Experimental Setup}
\noindent {\bf Datasets.}
We conduct the experiments on two commonly-used datasets: Math23K~\cite{wang-etal-2017-deep-neural} and MAWPS~\cite{koncel-kedziorski_mawps:_2016}. Math23K is a large-scale Chinese dataset that contains 23,162 math word problems and their corresponding expression solutions. MAWPS is a English dataset containing 2,373 problems. All the problems are one-unknown-variable linear problems and can be solved with a single expression.

\vspace{2mm}
\noindent {\bf Baselines.}
We compare our model with the following baselines including the state-of-the-art models: DNS~\cite{wang-etal-2017-deep-neural} uses a vanilla Seq2Seq model to generate expressions. Math-EN~\cite{wang_translating_2018} uses the equation normalization to avoid equation duplication problem. T-RNN~\cite{wang_template-based_2019} applies recursive neural networks to model the tree structures of expressions. S-Aligned~\cite{chiang_semantically-aligned_2019} tracks the semantic meanings of operands with a stack during decoding. Group-ATT~\cite{li_modeling_2019} leverages the attention mechanism to enrich problem representation. Both AST-Dec~\cite{liu_tree-structured_2019} and GTS~\cite{xie_goal-driven_2019} develop a tree-based decoder to generate expressions. Graph2Tree~\cite{zhang_graph--tree_2020} proposes to build a quantity cell graph and a comparison graph to better capture the quantity relationships of the problem. Multi-E/D~\cite{shen_solving_2020} is an ensemble model which combines multiple encoders and decoders. 

\vspace{2mm}
\noindent {\bf Implementation Details.}
We use the PyTorch\footnote{\url{https://pytorch.org/}} implementations and pre-trained language models provided by the Transformers library\footnote{\url{https://github.com/huggingface/transformers}}. Since the Math23K dataset is a Chinese dataset and officially released BART is only for English, we switch to mBART25~\cite{liu_multilingual_2020}, which is a multilingual BART for 25 languages including Chinese. For the MAWPS dataset, we also use mBART25. We optimize our model with AdamW~\cite{DBLP:conf/iclr/LoshchilovH19}. The training hyperparameters are set as follows. We set the batch size to 128, the learning rate to 5e-5 and the warm-up ratio to 0.1. The weight decay is set to 0.01. The number of epochs $M$ for fine-tuning and multi-task training are set to 50. We set beam size $K$ to 10 in beam search and expression bank size to 20 unless otherwise stated. All experiments are carried out on NVIDIA Tesla V100. We use 8 GPUs for training and 1 for testing. For our proposed framework, the training time is 1.5 hours for one epoch and testing time is 15 minutes for the whole test set.

\vspace{2mm}
\noindent {\bf Evaluation Metric.} Both MAWPS and Math23K are evaluated with a metric of ``solution accuracy'', that is, the expression is considered as correct if it induces the same number as the ground-truth. For the Math23K dataset, some baselines are evaluated using the public available test set while others use the results of 5-fold cross-validation. We report our results on both settings. For the MAWPS dataset, models are evaluated with 5-fold cross-validation.

\subsection{Results and Analysis}
Evaluation results of our model and baselines are summarized in Table \ref{tab:main_result}. We observe that: (1) direct fine-tuning of mBART already outperforms the state-of-the-art models on Math23K, which shows the powerful generation ability of mBART. (2) on MAWPS, mBART outperforms most Seq2Seq baselines but is worse than GTS and Graph2Tree. These two models leverage tree structure of expressions during decoding which is critical for math word problem solving. We believe that pre-trained language models would achieve a better performance if combined with structure information, and we leave it as a future work\footnote{One may think that the sequence decoder might not always generate valid expressions. However, we check all expressions generated by mBART and find that 99.9\% are valid.}. (3) \method\ framework further improves mBART and achieves new state-of-the-art results. In particular, \method\ outperforms mBART baselines by more than {\bf 4\%} in all the evaluation settings and also outperforms the previous best models by {\bf 7\%} on Math23K$^\dag$, {\bf 7.4\%} on 5-fold cross-validation Math23K$^\ddag$. The improvement over pre-trained mBART demonstrates the effectiveness of our multi-task training framework.

\begin{table}[h]
    \centering
    \resizebox{0.48\textwidth}{!}{
    \begin{tabular}{l|ccc}
    \hline
       Model  & Math23K$^\dag$ & Math23K$^\ddag$ & MAWPS$^\ddag$ \\
    \hline
       DNS   & - & 58.1& 59.5 \\ 
       Math-EN & 66.7 & - & 69.2 \\
       T-RNN  & 66.9 & - & 66.8\\
       S-Aligned & - & 65.8 & -\\
       Group-ATT & 69.5 & 66.9& 76.1\\
       AST-Dec & 69.0 & -& -\\
       GTS & 75.6& 74.3 & 82.6\\
       Graph2Tree & 77.4& 75.5& 83.7\\
       Multi-E/D & 78.4& 76.9 & -\\
       \hdashline
       mBART  & 80.8 & 80.0& 80.1\\
       Generate \& Rank & {\bf 85.4} & {\bf 84.3}& {\bf 84.0} \\
    \hline
\end{tabular}}
    \caption{Solution accuracy on MAWPS and Math23K. $\dag$ refers to the result of test set and $\ddag$ denotes the result of 5-fold cross-validation. ``-'' means that the results are not reported in the original papers.} 
    \label{tab:main_result}
\end{table}

\subsection{Ablation Study and Model Analysis}

To better understand our model, we further conduct ablation study on Math23K to show how the proposed components affect performance.


\subsubsection{Effect of Joint Training}
To investigate the effect of joint training, we introduce the baseline of two-stage training (i.e., w/o Joint), which means we first train the generator, then train the ranker, and the modules are trained independently. We also study the effect of joint training on generation and perform comparison between mBART and our generator (i.e., w/o Ranker).  The results are listed in Table \ref{tab:joint}. We can see that the joint training brings 2.2\% improvement compared with the two-stage training and 2.6\% for the generator compared with the mBART trained alone, suggesting that the joint training of generator and ranker benefits each other. Besides, the joint training is more space efficient since we only need to save one unified model rather than two.

\begin{table}[h]
    \centering
    \begin{tabular}{l|c}
        Model & Acc \\
    \hline
        \method & {\bf 85.4} \\
         w/o Joint & 83.2 \\
         w/o Ranker & 83.4 \\
         w/o both (mBART) & 80.8
    \end{tabular}
    \caption{Effect of joint training.}
    \label{tab:joint}
\end{table}


\subsubsection{Effect of Expression Bank Strategy}
We investigate the effect of different strategies to construct the expression bank. Here we choose a random sampling strategy as our baseline, where the set of expressions that appeared in the training data is sampled as the expression bank. We evaluate different strategies with and without online updating and summarize the results in Table \ref{tab:negative_sample}. 

\begin{table}[h]
    \centering
    \resizebox{0.45\textwidth}{!}{
    \begin{tabular}{l|c|c}
        Strategy &  Online & w/o Online \\
    \hline
        Random Sample & 75.2 & 69.7  \\
    \hdashline
        Model & 84.2& 83.2 \\
        Model+Tree & {\bf 85.4} & 83.1\\
    \end{tabular}}
    \caption{Accuracy for different expression bank strategies. The expression bank size is 20 for all settings.}
    \label{tab:negative_sample}
\end{table}


We can see that our strategies outperform the random sampling strategy. Since the ground-truth can not be accessed during model inference, we cannot use the tree-based disturbance to generate candidate expressions as in the training phase. This discrepancy between training and inference leads to poor performance if we only use tree-based disturbance to construct the expression bank. However, combining the  tree-based disturbance and model-based generation strategies, we can obtain better results than the only model-based generation, which gives evidence that the tree-based disturbance contains some informative examples that the generator does not cover and it is possible to improve the performance based on the human knowledge of math expression. 

We can also see that strategies have a performance drop without online updating. We conjecture that without online updating the ranker may tend to memorize existing negative expressions thus generalize poorly on new problems. As for strategies with model-based generation, there is another possible reason: the generator keeps updating during multi-task training, so the previously generated expressions are no longer good samples of the current model, and newly generated expressions are more informative. To summarize, both strategies of constructing the expressions bank and online updating play an important role in the success of the ranker.

\subsubsection{Impact of Expression Bank Size}
\begin{figure}[th]
    \centering
    \includegraphics[width=0.5\textwidth]{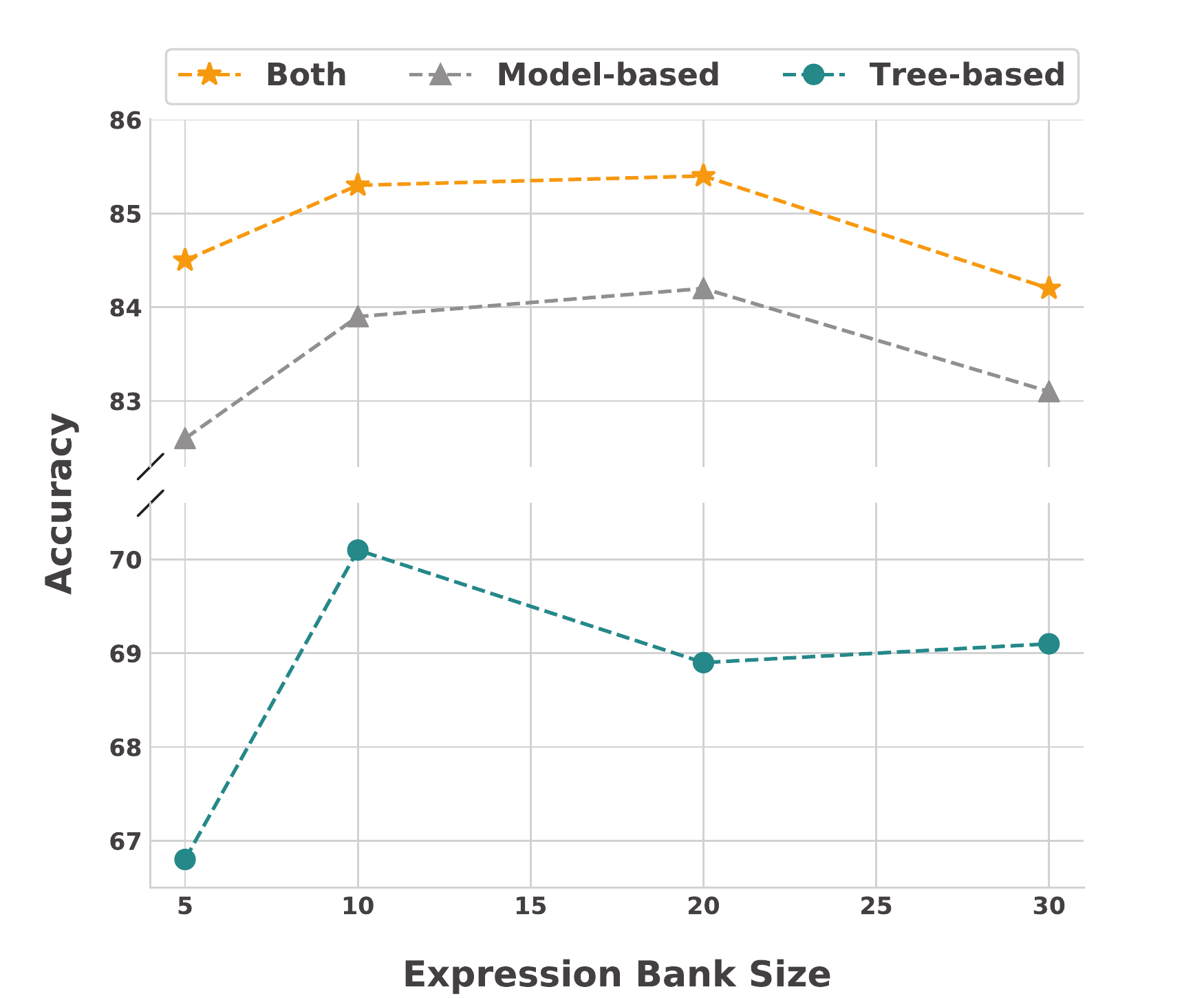}
    \caption{Accuracy with different expression bank sizes from 5 to 30.}
    \label{fig:bank_size}
\end{figure}

We further analyze the impact of expression bank size on the ranker and results are shown in Figure \ref{fig:bank_size}. If the model-based generation is used, performance reaches the best at expression bank size 20. This suggests that the expression bank size should not be too small nor too large. One possible reason may be that the generated expressions cannot cover possible mistakes when the expression bank is too small, and when the expression bank is too large, low-quality expressions may be generated and hinder ranker training. Tree-based disturbance has a similar trend and the best bank size is 10.

\subsubsection{Model Analysis}
In Table \ref{tab:error}, we list how the model accuracy changes with respect to the number of operations in expressions. We do not discuss the case of 6 operators since it has too few examples and high variance. For expressions less than 6 operators, all models perform worse when the expression gets longer. This is as expected since longer expressions require more steps of reasoning and have less data to train. In addition, we also observe that \method\ training has larger improvement over fine-tuned mBART on longer expressions. This implies that our model is more suitable to handle complex problems and expressions.

\begin{table}[t]
    \centering
    \resizebox{0.48\textwidth}{!}{
    \begin{tabular}{c|c|c|c|c|c}
        \#Op & Pro& AST-Dec&G2T& mBART& \method  \\
    \hline
        1 & 17.3 & 82.7 & 85.5 & 90.2& 90.8 (+0.6)\\
        2 &  52.2 & 74.5 & 83.7 & 88.1&90.2 (+2.1)\\
        3 & 19.1 & 59.9 & 71.7 & 71.2&79.1 (+7.9)\\
        4 & 6.6 & 42.4 & 51.5 & 53.0 &63.6 (+10.6)\\
        5 & 3.4 & 44.1 & 38.2 & 41.2&58.8 (+17.6)\\
        6 & 0.9 & 55.6 & 55.6 & 55.6 & 88.8 (+33.2)\\
    \end{tabular}}
    \caption{Accuracy for increasing length of expressions. \#Op is the number of operations in expressions. Pro denotes proportion of expressions with different lengths.}
    \label{tab:error}
\end{table}

Following \citet{liu_tree-structured_2019}, we also examine the performance of our model in different domains. The domain of each problem is defined by whether it contains any keywords of this domain and we use the same keyword list as \citet{liu_tree-structured_2019}. Table \ref{tab:domain} shows the results. We observe the similar pattern that the fine-tuned mBART has limitations in geometry which requires external knowledge such as formulas for the circumference and area of a circle. Interestingly, our proposed model mainly improves on these domains. This suggests that the ranking task may be a better choice to learn and use mathematical knowledge than generating.
\begin{table}[h]
    \centering
    \resizebox{0.47\textwidth}{!}{
    \begin{tabular}{l|c|c|c}
        Domain & Pro& mBART & \method  \\
        \hline
         Distance \& Speed& 11.8 & 83.9& 83.9 \\
         Tracing& 2.7 & 85.2& 85.2\\
         Engineering& 5.8 & 86.2& {\bf 87.9}\\
         Interval& 0.6 & 66.7& 66.7 \\
         Circle Geometry& 1.9 & 73.7 & {\bf 78.9}\\
         Plane Geometry& 1.2 & 75.0 & {\bf 83.3}\\
         Profit& 1.1 & 72.7 & 72.7\\
         Solid Geometry& 1.6 & 81.3 & {\bf 87.5}\\
         Interest Rate&0.9 & 100.0 &100.0\\
         Production & 0.4 & 100.0 & 100.0
    \end{tabular}}
    \caption{Accuracy for different problem domains. Pro denotes the proportion of each domain in the test data. Note that the sum of proportion is not 100\% since there are problems not belonging to any specified domain.}
    \label{tab:domain}
\end{table}

\section{Related Work}

\subsection{Math Word Problem}
\noindent {\bf Rule-based methods.}
Early approaches on math word problems mainly craft rules and templates for pattern matching~\cite{bobrow_natural_1964, slagle_experiments_1965, fletcher_understanding_1985, bakman_robust_2007}. These methods rely heavily on manual design and can only solve a limited scope of problems. 

\vspace{2mm}
\noindent {\bf Parsing-based methods.}
Later on, researchers use statistical methods to solve MWP and achieve a great performance improvement. One line of research focuses on semantic parsing, which leverages traditional machine learning techniques to identify entities, quantities, and operators from the problem text. \citet{roy_reasoning_2015} proposes three types of classifiers to identify different elements of problems. ARIS~\cite{hosseini_learning_2014} splits the problem into fragments and updates a logic template named \textit{state} by verb categorization. Other works~\cite{sundaram_natural_2015,mitra_learning_2016,liang_tag-based_2016} follow a similar process with different templates and annotations.

\vspace{2mm}
\noindent {\bf Two-stage methods.}
Another research line first obtains an expression template then maps numbers to the template slots. \citet{kushman_learning_2014} train a classifier to select from a set of pre-defined templates. \citet{roy_solving_2015} propose to construct candidate expressions in a bottom-up manner and train a global scoring function to guide the beam search process. ALGES~\cite{koncel-kedziorski_parsing_2015} converts the process of searching valid expressions to an integer linear programming problem and adopts a different scoring function. UnitDep~\cite{roy_unit_2017} proposes Unit Dependency Graph to enhance the scoring function.

\vspace{2mm}
\noindent {\bf Deep learning methods.}
Recently, deep learning models have become prevailing methods for math word problems. DNS~\cite{wang-etal-2017-deep-neural} is the first to apply vanilla RNN-based models to MWP. Math-EN~\cite{wang_translating_2018} introduces equation normalization and compares three Seq2Seq models on MWP solving. Group-ATT~\cite{li_modeling_2019} uses multi-head attention to capture different aspects of features. Some works also leverage tree structures and graph information to improve performance ~\cite{wang_template-based_2019,chiang_semantically-aligned_2019,liu_tree-structured_2019,xie_goal-driven_2019, zhang_graph--tree_2020}. \citet{shen_solving_2020} propose a model of multi-encoders and multi-decoders. 

\subsection{Pre-trained Language Model}
Pre-trained language models have obtained state-of-the-art results in many NLP benchmarks~\cite{wang_glue:_2018, wang_superglue:_2019}. These models are usually based on Transformer layers~\cite{Vaswani_attention} and trained on large corpus with self-supervised tasks. According to their architectures, pre-trained language models can be categorized into three types: encoder-only, decoder-only and encoder-decoder models. BERT~\cite{devlin_bert:_2019} is an encoder-only model which firstly proposes masked token prediction and next sentence prediction to train a language representation model. Following this, many other models are proposed like RoBERTa~\cite{liu_roberta:_2019} and SpanBERT~\cite{joshi_spanbert:_2020}. Decoder-only models are typically auto-regressive models trained to estimate the probability distribution of a text corpus, including GPT2~\cite{radford2019language}, GPT3~\cite{brown_language_2020} and XLNet~\cite{NEURIPS2019_dc6a7e65}. Encoder-decoder models like BART~\cite{lewis_bart:_2020} and T5~\cite{raffel_exploring_2020} use the encoder-decoder architecture and are trained on sequence-to-sequence tasks such as text denoising and translation. 
\section{Conclusion and Future Work}
We propose \method, a new multi-task framework for math word problems. Specifically, our model has a generator and a ranker which enhance each other with joint training. We also use tree-based disturbance and online update to further improve the performance. The experimental results on the benchmark show that our work consistently outperforms baselines in all datasets. In future work, we will explore the generation and ranking framework to other tasks like summarization and translation.
\section*{Acknowledgements}
This paper is partially supported by National Key Research and Development Program of China with Grant No. 2018AAA0101900/2018AAA0101902 as well as the National Natural Science Foundation of China (NSFC Grant No. 62106008 and No. 61772039). 

\bibliography{anthology,custom}
\bibliographystyle{acl_natbib}

\end{document}